\pdfoutput=1

\documentclass[11pt]{article}

\usepackage[final]{acl}

\usepackage{times}
\usepackage{latexsym}

\usepackage[T1]{fontenc}

\usepackage[utf8]{inputenc}

\usepackage{microtype}

\usepackage{inconsolata}

\usepackage{graphicx}

\usepackage{enumitem}
\usepackage{amsfonts}
\usepackage{amsmath}
\usepackage{multirow}
\usepackage{booktabs}
\usepackage[table]{xcolor}
\usepackage{makecell} 
\usepackage{amssymb}
\usepackage{pifont}
\usepackage{bm}

\title{Static or Dynamic: Towards Query-Adaptive Token Selection for \\ Video Question Answering}

\author{Yumeng Shi \quad Quanyu Long \quad Wenya Wang \\
       Nanyang Technological University \\
       \texttt{yumeng001@e.ntu.edu.sg} \quad \texttt{quanyu001@e.ntu.edu.sg} \quad \texttt{wangwy@ntu.edu.sg}}

\begin{document}
\maketitle
\begin{abstract}
Video question answering benefits from the rich information in videos, enabling various applications. However, the large volume of tokens generated from long videos presents challenges to memory efficiency and model performance. To alleviate this, existing works propose to compress video inputs, but often overlook the varying importance of static and dynamic information across different queries, leading to inefficient token usage within limited budgets. We propose a novel token selection strategy, \textsc{explore-then-select}, that adaptively adjusts static and dynamic information based on question requirements. Our framework first explores different token allocations between key frames, which preserve spatial details, and delta frames, which capture temporal changes. Then it employs a query-aware attention-based metric to select the optimal token combination without model updates. Our framework is plug-and-play and can be seamlessly integrated within diverse video language models. Extensive experiments show that our method achieves significant performance improvements (up to 5.8\%) on multiple video question answering benchmarks. Our code is available at \href{https://github.com/ANDgate99/Explore-Then-Select}{\textit{https://github.com/ANDgate99/Explore-Then-Select}}.

\end{abstract}

\section{Introduction}
Video Question Answering (VideoQA) has broad applications across various fields~\cite{mogrovejo2024question, zhang-etal-2024-simple}. Compared to text, videos provide more intuitive and dynamic information, delivering richer context and details by combining visual and temporal elements. Current research primarily leverages powerful large language models to build video language models (VideoLMs)~\cite{lin2023video, zhang2024longva}, significantly enhancing AI performance in VideoQA tasks. However, the extensive visual information in long videos leads to a dramatic increase in token counts. For instance, if one frame generates 196 tokens~\cite{li2024llava}, a 5-minute video sampled at 1 fps would produce nearly 60,000 tokens, posing significant challenges to memory requirements and model capabilities. 

\begin{figure}[t]
  \includegraphics[width=\columnwidth]{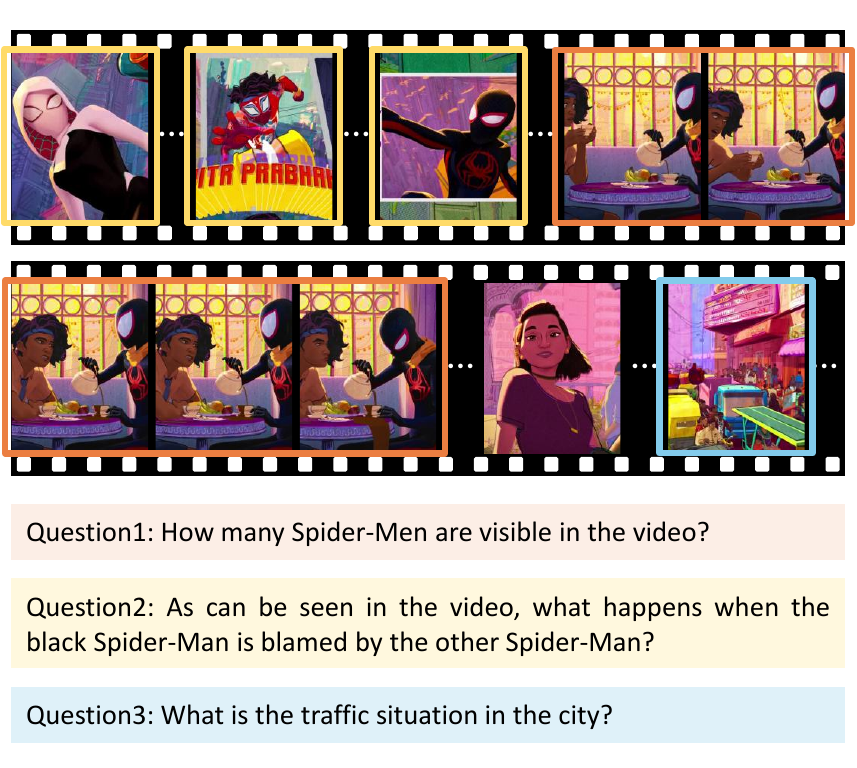}
  \caption{Different question types vary in their dependence on static and dynamic information in videos. For example, Question 2 relies on fine-grained dynamic information, while Question 1 and 3 only require key frames. The frames needed to answer the questions are highlighted with corresponding colored boxes.}
  \label{fig:intro}
\end{figure}

Given the strict token limitations in practical VideoLM deployments, effectively representing essential video information requires a careful allocation between static and dynamic content. Static information, which refers to the visual content within individual frames, is crucial for questions like object recognition, where spatial details dominate. In contrast, dynamic information captures temporal changes and motion patterns across consecutive frames, which are essential for understanding actions or events. Figure~\ref{fig:intro} illustrates different types of questions, which vary in their reliance on static and dynamic information. Considering these varying dependencies, the challenge lies in optimizing the allocation of limited tokens to preserve the most relevant aspects of both static and dynamic information, depending on specific question requirements. Although existing studies~\cite{shen2024longvu, nie2024slowfocus} have explored token compression through changing frame sampling rates or intra-frame downsampling, they fail to address the varying dependencies on static and dynamic information across different question types.

To achieve an effective allocation between static and dynamic information in visual token compression, we propose a novel token selection strategy, \textsc{Explore-then-Select}, that adaptively aligns visual tokens with textual queries under a limited token budget. Unlike previous approaches that rely on fixed rules, our strategy autonomously and adaptively combines static and dynamic content based on the nature of the questions (e.g., action description, event sequence, or object recognition), ensuring more precise responses to diverse queries.

Specifically, we categorize video frames into key and delta frames. Key frames are fully retained to preserve essential spatial details, such as objects, while delta frames are sparsely processed, keeping only a subset of tokens to capture important temporal changes. To optimize token allocation between these two types of frames, \textsc{Explore-then-Select} uses a two-stage process. In the \textbf{exploration} stage, we construct a search space comprising various combinations of key and delta frames, each yielding a token subsequence of constrained length. By adjusting the proportion of key and delta frames, we can prioritize either static details or dynamic changes based on the question requirements. In the \textbf{selection} stage, we evaluate each combination using a query-aware metric derived from the shallow attention layers of VideoLMs. This metric quantifies the alignment between the query and visual tokens, enabling us to select the optimal combination to answer the question.

Notably, our framework is training-free, as neither the exploration nor selection processes require model updates. Leveraging its seamless integration with diverse VideoLMs, we demonstrate the effectiveness of our approach on two widely recognized VideoLMs across multiple benchmarks for both long and short videos. Using our framework, models can achieve improvements of up to 5.8\%. Our key contributions are summarized as follows:
\begin{itemize}
    \item Building on the observation that questions rely differently on static and dynamic video information, we propose a novel \textsc{Explore-then-Select} framework to adaptively and effectively select visual tokens reflecting the optimal balance of static and dynamic information under limited token budgets.
    \item To address static and dynamic information needs, we design an effective search space of key-delta frame combinations. During the selection phase, we employ a query-aware approach, leveraging an attention-based metric to adaptively evaluate candidates and select the optimal combination for each question.
    \item We conduct extensive experiments on both long and short video benchmarks, demonstrating the effectiveness of our method. Thanks to its plug-and-play design, our approach generalizes well across different models without extra fine-tuning and enables direct control over the token budget for flexible adaptation to resource constraints.
\end{itemize}

\section{Related Work}
\newcommand{\cmark}{\ding{51}}
\newcommand{\xmark}{\ding{55}}
\newcommand{\gc}{\cellcolor{gray!17}}
\newcommand{\tb}{\textbf}

\subsection{Video Language Models}
Significant progress has been made in video language model research based on LLMs. These models can be primarily classified into two types: general-purpose vision language models~\cite{team2024gemini, chen2024internvl, openai2024gpt4o, yao2024minicpm, ye2023mplug} and specialized video language models~\cite{lin2023video, zhang2024video, li2024llama, damonlpsg2025videollama3, kangaroogroup}. Among the former, LLaVA-OneVision~\cite{li2024llava} unifies image and video tasks, while Qwen2-VL~\cite{wang2024qwen2} introduces dynamic resolution support and three-dimensional positional encoding for enhanced visual feature capture. Among specialized models, VideoChat~\cite{li2023videochat} targets deep video understanding and interaction, and LongVA~\cite{zhang2024longva} extends the context length of language models, transferring their advantages in long-text processing to the video domain.

\begin{table}[t]
    \centering
    \fontsize{9}{11}\selectfont
    \begin{tabular}{lccc}
        \toprule
        \multirow{2}{*}{\textbf{Method}} & \textbf{Pre-} &  \makebox[0.65cm]{\textbf{Training-}} & \makebox[0.65cm]{\textbf{Video-}}\\
         & \makebox[0.65cm]{\textbf{Input}} & \makebox[0.65cm]{\textbf{Free}} & \makebox[0.65cm]{\textbf{Specific}}\\
        \midrule
        \makebox[4cm][l]{FastV~\cite{chen2024image}}          & \xmark & \cmark  & \xmark \\
        \makebox[4cm][l]{ZipVL~\cite{he2024zipvl}}            & \xmark & \cmark  & \xmark \\
        \makebox[4cm][l]{FrameFusion~\cite{fu2024framefusion}} & \xmark & \cmark  & \cmark \\
        \makebox[4cm][l]{TokenPacker~\cite{li2024tokenpacker}} & \cmark & \xmark  & \xmark \\
        \makebox[4cm][l]{VideoStreaming~\cite{qian2025streaming}}   & \cmark & \xmark  & \cmark \\
        \makebox[4cm][l]{SlowFocus~\cite{nie2024slowfocus}}    & \cmark & \xmark  & \cmark \\
        \makebox[4cm][l]{LongVU~\cite{shen2024longvu}}         & \cmark & \xmark  & \cmark \\
        \makebox[4cm][l]{\gc Ours}                             & \gc \cmark & \gc \cmark  & \gc \cmark \\
        \bottomrule
    \end{tabular}
    \caption{Feature comparison with existing methods. ``Pre-Input'' refers to methods that reduce tokens before feeding them into large language models, while ``Video-Specific'' denotes methods that leverage the unique characteristics of video data.}
    \label{tab:method}
\end{table}

\subsection{Visual Token Compression}
Some studies~\cite{bolya2022token} focus on compressing visual tokens in vision encoders. For example, RLT~\cite{choudhury2024don} effectively reduces the number of tokens by replacing repeated patches in videos with a single patch. Other works~\cite{li2024tokenpacker, qian2025streaming, shen2024longvu, lan2024vidcompress} introduce dedicated modules for token compression, such as BLIP-2~\cite{li2023blip}, which uses a Q-Former module with learnable queries to generate compact semantic representations. Additionally, inspired by KV cache compression in long text processing~\cite{zhang2023h2o}, some methods apply similar strategies to visual tokens~\cite{he2024zipvl, chen2024image, fu2024framefusion}. These methods optimize token usage efficiency by setting thresholds based on specific metrics to prune visual tokens.

Table~\ref{tab:method} compares existing methods, noting that training-free approaches mainly compress tokens within the KV cache, reducing FLOPs but failing to address the issue of excessive token input to large language models. In contrast, methods that reduce tokens in advance typically require training. This paper introduces a novel pre-input, training-free framework for more effective compression, balancing query-aware static and dynamic information.

\section{Preliminary}
In this section, we outline the common inference pipeline of VideoLMs as the setup for our approach. It consists of three key steps, including video frame sampling, visual encoding and embedding, and multimodal inference.

\begin{itemize}[leftmargin=0pt, itemindent=0pt]
\item[] \textbf{Video Frame Sampling.} Given an input video, $N$ frames are uniformly sampled to form a representation $\bm{V} \in \mathbb{R}^{N \times C \times H_v \times W_v}$, where $C = 3$ denotes the RGB channels, and $H_v$ and $W_v$ represent the height and width of each frame, respectively.


\item[] \textbf{Visual Encoding and Embedding.} The sampled frames are divided into non-overlapping spatiotemporal patches, which are processed by a vision encoder to extract spatiotemporal features. These features are then projected into the language model's token space via a linear projection, resulting in visual token embeddings $\bm{F} \in \mathbb{R}^{T \times H \times W \times D}$, where $T$ represents the temporal resolution (typically equal to $N$ unless temporal downsampling is applied), $H$ and $W$ denote the spatial resolutions, and $D$ is the token embedding dimension.
 

\item[] \textbf{Multimodal Processing.} The visual token embeddings $\bm{F}$ are then flattened into a sequence $\bm{T}_v \in \mathbb{R}^{L \times D}$, where $L=T\times H\times W$ is the sequence length. The sequence $\bm{T}_v$, instruction embeddings $\bm{T}_i$, and query embeddings $\bm{T}_q$ are concatenated into a unified input $\bm{T} = [ \bm{T}_i, \bm{T}_v, \bm{T}_q]$, where $[\cdot]$ denotes token concatenation. Finally, the VideoLM processes the unified input sequence $\bm{T}$ to generate a textual response to the question.

\end{itemize}

\begin{figure*}[t]
  \includegraphics[width=\textwidth]{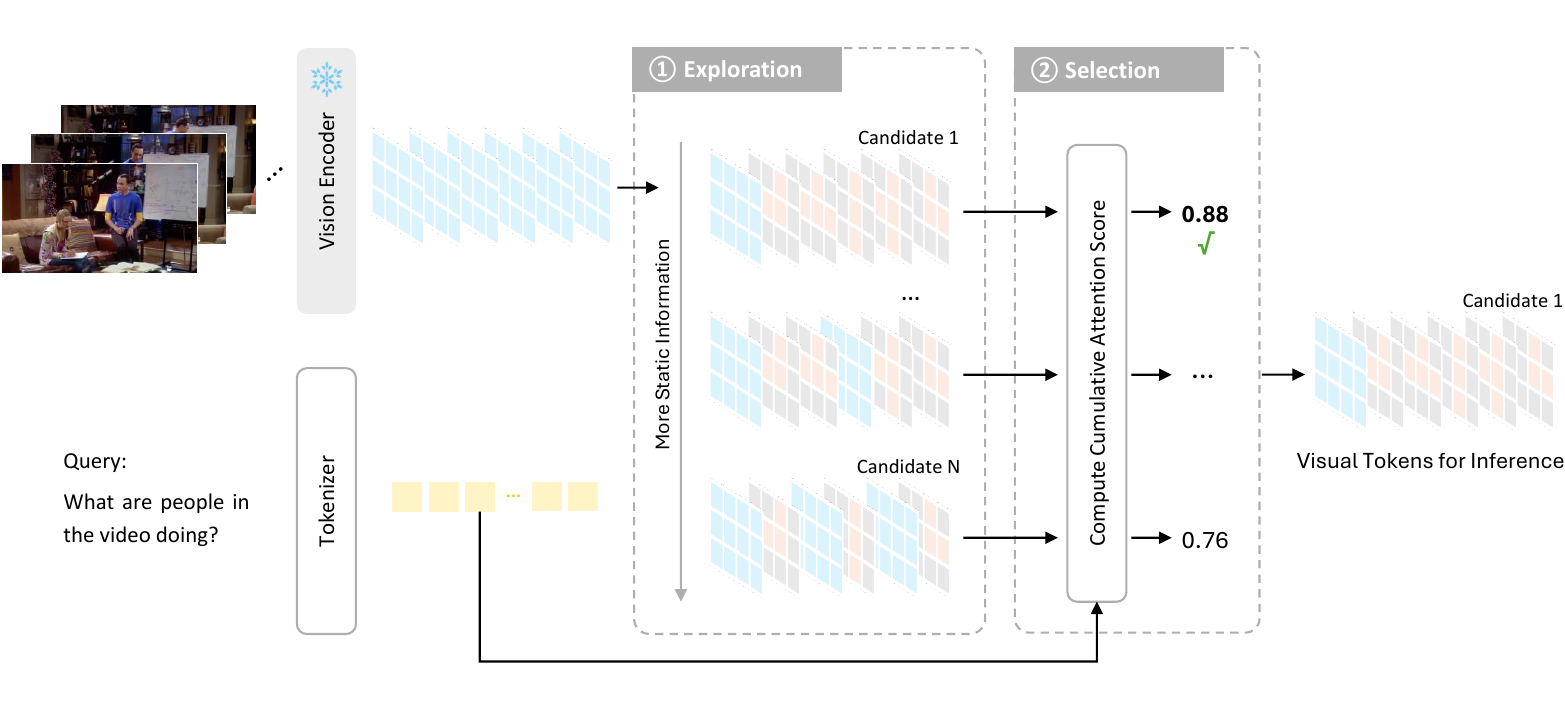}
  \caption{Overview of our \textsc{Explore-then-Select} framework for token selection. During the exploration stage, multiple subsequences are generated from different combinations of key and delta-frame tokens. In the selection stage, these subsequences are evaluated using query-aware metrics computed from shallow attention layers, and the optimal subsequence is chosen as input to the LLM.}
  \label{fig:overview}
\end{figure*}

\section{Method}
\subsection{Problem Definition}
Due to GPU memory and model capability constraints, the number of visual tokens processed during inference is capped at $L_b$. The fixed token budget limits frame sampling to a reduced number of frames, resulting in a significant loss of rich visual information, particularly in long videos.

In this work, we aim to sample more frames to expand the amount of information we can capture, which generates an excessive number of tokens, leading to a sequence length $L \gg L_b$. Then we compress the tokens to meet the token budget, enabling more effective utilization of rich visual information within the limited length.

To meet our goal, we propose a token-efficient framework that automatically and adaptively selects a limited yet informative set of visual tokens by leveraging the textual query's relevance to both static and dynamic visual information. Our method emphasizes balancing these two types of information, ensuring that the selected tokens maximize their alignment with the query while maintaining memory efficiency.

\subsection{Framework Overview}
We adopt an \textsc{Explore-then-Select} framework, as illustrated in Figure~\ref{fig:overview}. In the token exploration stage (Section~\ref{sec:method1}), we construct a search space of $n$ visual token subsequences, each of length $L_b$, where every visual token subsequence reflects a distinct balance of static and dynamic information. In the token selection stage (Section~\ref{sec:method2}), we identify the optimal sequence that best aligns with the query requirements. Details will be discussed below.



\subsection{Exploration: Search Space Design} \label{sec:method1}
In this section, we describe the generation of $n$ token subsequences, each of length $L_b$, from a token sequence of length $L$. To balance static and dynamic information in videos according to query requirements, we classify frames into key and delta frames. Note that, due to temporal downsampling in some models, ``frames'' here refer to visual token embeddings $\bm{F}$, and the total number of frames is $T$. Based on whether the tokens in a subsequence originate from key or delta frames, we divide them into two subsets: key-frame tokens $\mathcal{T}_{\text{key}}$ and delta-frame tokens $\mathcal{T}_{\text{delta}}$.

\begin{figure}[t]
  \includegraphics[width=\columnwidth]{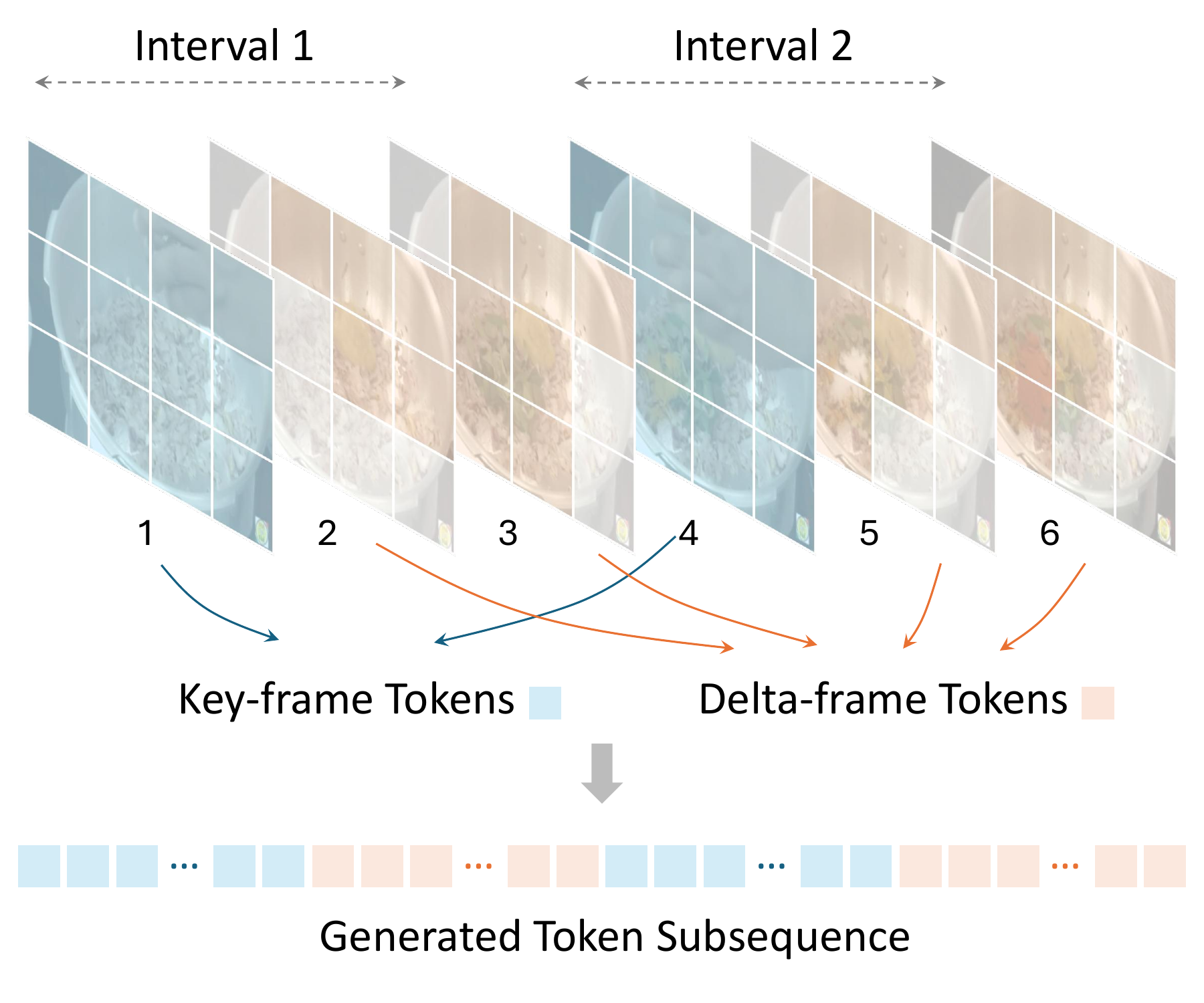}
  \caption{An example of token subsequence generation with 6 total frames and 2 key frames.}
  \label{fig:method1}
\end{figure}

\begin{itemize}[leftmargin=0pt, itemindent=0pt]
\parindent 1em
\item[] \textbf{Key-frame Token.} The key-frame tokens are extracted from the key frames. Assuming $N_s$ key frames are selected in the video, we select them uniformly from $\bm{F}$. The temporal indices of these frames are:
\begin{equation}
\label{eq:1}
\begin{gathered}
    \mathcal{I} = \left\{ \left\lfloor \frac{k T}{N_s} \right\rfloor + 1\mid k = 0, 1, \dots, N_s - 1 \right\},
\end{gathered}
\end{equation}
where the first frame is always selected as a key frame. All tokens from these key frames are retained to form $\mathcal{T}_{\text{key}}$:
\begin{equation}
\label{eq:2}
     \{ \bm{F}^{i,h,w} \mid i \in \mathcal{I}, h\in [1,H], w\in [1,W] \},
\end{equation}
where $\bm{F}^{i,h,w} \in \mathbb{R}^D$ represents the token embedding at the $i$-th frame and spatial location $(h, w)$ in $\bm{F}$. Hence, the total number of key-frame tokens is $|\mathcal{T}_{\text{key}}| = N_s \times H \times W$.

\item[] \textbf{Delta-frame Token.} 
As illustrated in Figure~\ref{fig:method1}, the key frames partition the entire sampled frame sequence into $N_s$ intervals. 
The frames between consecutive key frames within each interval are defined as delta frames, and delta-frame tokens $\mathcal{T}_{\text{delta}}$ are extracted from them to capture dynamic information relative to the preceding key frames. 
Given a subsequence length of $L_b$, the total number of delta-frame tokens is 
$|\mathcal{T}_{\text{delta}}| = L_b - |\mathcal{T}_{\text{key}}|$. 
These tokens are uniformly distributed across the intervals, such that the number of delta-frame tokens selected from the $i$-th interval is 
$|\mathcal{T}_{\text{delta},i}| = \lfloor |\mathcal{T}_{\text{delta}}| / N_s \rfloor$. \\
\indent Inspired by video coding techniques, to retain as much dynamic information as possible, we select tokens from each interval that exhibit the largest differences compared to the corresponding tokens in the preceding key frame. We first define the token difference metric based on the cosine similarity between two token embeddings:
\begin{equation}
\label{eq:3}
    \mathcal{D}(\bm{f}_i, \bm{f}_j) = 1-\frac{\bm{f}_i \cdot \bm{f}_j}{\|\bm{f}_i\| \|\bm{f}_j\|}.
\end{equation}
\indent This metric increases as the two embeddings become more dissimilar. \\
\indent For interval $i$, we compute the difference between each token in the frames within the interval and the token at the corresponding spatial position in the preceding key frame. Specifically, we define:
\begin{equation}
\label{eq:new}
\begin{gathered}
    \Delta_i(j, h, w) = \mathcal{D}(\bm{F}_i^{0,h,w}, \bm{F}_i^{j,h,w}) \\  
    j \in [1, T_i], h \in [1,H], w \in [1,W],
\end{gathered}
\end{equation}
where $\bm{F}_{i}^{0,h,w}$ denotes the token embedding at spatial location $(h,w)$ in the preceding key frame of interval $i$, and $\bm{F}_{i}^{j,h,w}$ denotes the token embedding at the same spatial location in the $j$-th delta frame within the interval. Here, $T_i$ is the number of delta frames contained in interval $i$. \\
\indent We then select the delta-frame tokens corresponding to the $|\mathcal{T}_{\text{delta},i}|$ largest values of $\Delta_i(j,h,w)$, forming the set $\mathcal{T}_{\text{delta},i}$ as:
\begin{equation}
\label{eq:4}
\{ \bm{F}_i^{j,h,w} \mid (j,h,w) \in
\operatorname{arg\,Top}_{|\mathcal{T}_{\text{delta},i}|} \Delta_i(j,h,w) \}.
\end{equation}


\item[] \textbf{Token Subsequence Generation.} Then we merge $\mathcal{T}_{\text{key}}$ and $\mathcal{T}_{\text{delta}}$ according to their original order to obtain the $L_b$-long token subsequence $\bm{\hat{T}}_v$.

\end{itemize}

To generate $n$ candidate token subsequences, we vary the number of key frames $N_s$ from $1$ to $n$. A smaller $N_s$ results in more delta-frame tokens, thereby capturing more dynamic information within the same token budget. Conversely, a larger $N_s$ increases the number of key-frame tokens, preserving more static information. In this way, we can generate token subsequences with varying proportions of static and dynamic information to adapt to the requirements of different queries.

Notably, our frame division is inspired by the GOP structure in video codec~\cite{lee2006adaptive}, where I-frames encode full scene information and P/B-frames encode temporal differences. Similar to adjusting GOP sizes, varying the proportion of key and delta frames allows us to control the emphasis on static or dynamic cues.

\subsection{Selection: Quick Evaluation} \label{sec:method2}
After obtaining $n$ token subsequences of length $L_b$, we perform an evaluation and select the optimal subsequence based on the chosen metric. Previous studies have identified certain characteristics of visual tokens in attention mechanisms. For instance, \citet{chen2024image} shows that most visual tokens can be removed at the second layer without significant performance loss, and \citet{wan2024look} observes that visual tokens are generally less attended. Based on these findings, we consider that the attention mechanism at the second layer already provides meaningful clues of token importance. Besides, we hypothesize that higher cumulative attention scores on visual tokens indicate a better utilization of the visual information.

\begin{table*}[t]
    \centering
    \fontsize{9}{11}\selectfont
    \begin{tabular}{llcccccccc}
        \toprule
        \multirow{2}{*}{\textbf{Model}} & \multicolumn{3}{c}{\textbf{Settings}} & \textbf{EgoSchema} & \multicolumn{4}{c}{\textbf{VideoMME}} & \textbf{MLVU} \\
        & Method & Sample & Budget & & Short & Medium & Long & Overall \\
        \midrule
        VideoChat2
        & -	  & 16	 & -	      & 54.4 & 48.3	& 37.0 & 33.2 & 39.5 & -      \\
        LongVA
        & -   & 128  & -          & -    & 61.1 & 50.4 & 46.2 & 52.6 & -      \\
        mPLUG-Owl3
        & -   & 128  & -          & -    & 70.0 & 57.7 & 50.1 & 59.3 & -      \\
        LongVU
        & -   & 1fps & -          & 67.6 & -    & -    & -    & 60.6 & 65.4   \\
        \midrule
        \multirow{8}{*}{Qwen2-VL-7B} 
        & Original   & 64  & -    & 66.2 & 71.1 & 59.4 & 50.8 & 60.4 & 50.6 \\
        & Retrieval  & 256 & 64   & 63.6 & 71.0 & 61.3 & 52.2 & 61.5 & 49.4 \\
        & Similarity & 256 & 64   & 66.6 & 71.4 & 60.6 & 51.8 & 61.3 & 53.0 \\
        & \gc Ours   & \gc 256 & \gc 64   & \gc \tb{67.8} & \gc \tb{72.4} & \gc \tb{63.1} & \gc \tb{53.2} & \gc \tb{62.9} & \gc \tb{54.4} \\ 
        \cmidrule(lr){2-10}
        & Original   & 32  & -    & 64.7 & 68.9 & 55.2 & 48.7 & 57.6 & 46.8 \\
        & Retrieval  & 128 & 32   & 61.7 & 70.0 & 58.6 & 51.6 & 60.0 & 46.8 \\
        & Similarity & 128 & 32   & 65.6 & 70.1 & 58.7 & \tb{51.8} & 60.2 & 47.2 \\
        & \gc Ours   & \gc 128 & \gc 32   & \gc \tb{66.7} & \gc \tb{71.4} & \gc \tb{61.0} & \gc 51.7 & \gc \tb{61.4} & \gc \tb{52.2} \\
        \midrule
        \multirow{8}{*}{LLaVA-OneVision-7B} 
        & Original   & 64  & -    & 60.1 & 70.6 & 55.8 & 47.8 & 58.0 & 50.8 \\
        & Retrieval  & 256 & 64   & 57.7 & 64.0 & 53.4 & 47.0 & 54.8 & 44.6 \\
        & Similarity & 256 & 64   & 59.6 & 71.0 & 57.9 & 50.8 & 59.9 & 48.4 \\
        & \gc Ours   & \gc 256 & \gc 64   & \gc \tb{60.3} & \gc \tb{71.9} & \gc \tb{58.3} & \gc \tb{51.4} & \gc \tb{60.6} & \gc \tb{51.2} \\
        \cmidrule(lr){2-10}
        & Original   & 32  & -    & 60.4 & \tb{71.3} & 57.4 & 48.0 & 58.9 & 46.8\\
        & Retrieval  & 128 & 32   & 57.9 & 63.2 & 53.9 & 46.0 & 54.4 & 44.0\\
        & Similarity & 128 & 32   & 60.2 & 70.8 & 57.1 & 49.7 & 59.2 & 50.2 \\
        & \gc Ours   & \gc 128 & \gc 32   & \gc \tb{60.5} & \gc 70.2 & \gc \tb{58.0} & \gc \tb{51.6} & \gc \tb{59.9} & \gc \tb{51.0} \\
        \bottomrule
    \end{tabular}
    \caption{Results on long video benchmarks show that our method achieves significant improvements over the baselines, particularly on the advanced Qwen2-VL, with up to a 5.8\% gain on the VideoMME medium subset.}
    \label{tab:long}
\end{table*}

To enable quick evaluation, we compute the attention score matrix $\bm{S}$ at the second layer of the VideoLMs, using textual query tokens as the query input, and instruction and visual tokens as the key input:
\begin{equation}
\label{eq:5}
    \bm{Q} = \bm{W}_\text{Q}\bm{H}_q,
\end{equation}
\begin{equation}
\label{eq:6}
    \bm{K} = \bm{W}_\text{K} \, \text{concat}(\bm{H}_{i},\bm{H}_{v}),
\end{equation}
\begin{equation}
\label{eq:7}
    \bm{S} = \text{softmax}\left(\frac{\bm{QK}^\top}{\sqrt{d_{k}}}\right),
\end{equation}
where $\bm{H}_q$, $\bm{H}_i$, and $\bm{H}_v$ denote the hidden features of the textual query, instruction, and visual inputs. Here, $d_k$ is the dimension of key vectors in the attention mechanism. To quantify the attention allocated to the visual tokens, we compute the summed attention scores of the visual tokens. Specifically, to ensure comprehensive consideration of each text query token, we first extract the maximum values along the query dimension from $\bm{S}$, yielding an attention score vector $\bm{s}$ for each visual token. Then we sum the attention scores of the visual tokens:
\begin{equation}
\label{eq:8}
    \bm{s} = \max_{i} \bm{S}_{ij},
\end{equation}
\begin{equation}
\label{eq:9}
    s = \sum_{j=N_{i}}^{N_i+N_v} \bm{s}_j,
\end{equation}
where $\bm{S}_{ij}$ represents the attention score of the $i$-th query token to the $j$-th visual token, $N_i$ denotes the number of instruction tokens, and $N_v$ is the number of visual tokens. Finally, from the $n$ candidates, we select the input with the highest sum of visual token attention scores as the optimal input:
\begin{equation}
\label{eq:10}
    {\bm{\bar{T}}_v} = \underset{ m \in \{1, 2, \dots, n\}}{\arg\max}s^{m},
\end{equation}
where $s^{m}$ denotes the summed attention score for the $m$-th token subsequence.

\section{Experiments}

\subsection{Experimental Settings}

\begin{itemize}[leftmargin=0pt, itemindent=0pt]
\parindent 1em
\item[] \textbf{Benchmarks.} To comprehensively evaluate performance, we select benchmarks for both long and short videos. We use VideoMME, EgoSchema, and MLVU for long videos, and MSVD-QA and ActivityNet-QA for short videos.\\
\indent VideoMME~\cite{fu2024video} contains 900 videos (11 seconds to 1 hour) and 2,700 QA pairs. EgoSchema~\cite{mangalam2023egoschema} includes over 5,000 questions based on videos averaging 3 minutes in length. MLVU~\cite{MLVU} provides videos ranging from 3 minutes to 2 hours, with the test set containing over 500 QA pairs. MSVD-QA~\cite{xu2017video} includes 1,970 short clips (10 seconds on average), with a test split of approximately 13,000 questions. ActivityNet-QA~\cite{yu2019activitynet} provides 800 videos and 8,000 QA pairs in the test set.\\
\indent We adopt multiple-choice accuracy as the evaluation metric for VideoMME, EgoSchema, and MLVU, and employ GPT-4o mini~\cite{openai2024gpt4o} to score answers for the open-ended MSVD-QA and ActivityNet-QA.

\item[] \textbf{Baselines.} We validate our plug-and-play method on two representative models: Qwen2-VL~\cite{wang2024qwen2}, featuring dynamic resolution and multimodal rotary position embeddings, and LLaVA-OneVision~\cite{li2024llava}, supporting multiple tasks, both in their 7B versions. Results for Qwen2.5-VL~\cite{Qwen2.5-VL} are included in Appendix~\ref{appendix_qwen2.5}. \\
\indent As shown in Table~\ref{tab:method}, prior methods either compress only within the KV cache, leaving long input sequences unaddressed, or require training models, making direct comparison with our training-free approach unfair. Thus, we consider three baselines: 1) Original: uniform frame sampling within the token budget; 2) Retrieval: oversample frames, then prune based on cosine similarity between frame and query embeddings to fit the token limit; 3) Similarity: oversample frames, then prune based on cosine similarity between adjacent token embeddings. In practice, both ``Retrieval'' and ``Similarity'' strategies are commonly adopted in compression modules~\cite{qian2025streaming, song2024moviechat, he2024ma}. For reference, we also report results from several training-based video understanding methods~\cite{li2023videochat, zhang2024longva, ye2024mplug, shen2024longvu} in the first block of Table~\ref{tab:long}, though they are not directly comparable due to training cost differences. To further validate the advantages of our method, we include a comparison with our reproduced training-free LongVU in Appendix~\ref{app_longvu}.

\item[] \textbf{Implementation Details.} 
All experiments are run on two 40GB A100 GPUs or one 80GB A100 GPU. For multiple-choice questions, the model generates one token (three for MLVU), while for open-ended questions, outputs are limited to 30 tokens. The prompts used are detailed in Appendix~\ref{sec:appendix}. Sampling is disabled to ensure deterministic results.\\
\indent Note that video resolution affects the number of frame tokens generated by Qwen2-VL, making a fixed token budget yield varying frame counts across videos and complicating comparisons. To address this, we set a frame-based budget $T_b$, so the token limit is $L_b = T_b \times H \times W$, where $H \times W$ is the token count per frame. This approach streamlines implementation and ensures fair comparison. Besides, unless otherwise specified, the number of subsequences generated during the exploration stage is set to half of the frame budget $T_b$.

\end{itemize}
\subsection{Main Results}

\begin{itemize}[leftmargin=0pt, itemindent=0pt]
\parindent 1em
\item[] \textbf{Long Video Results.} Table~\ref{tab:long} shows results on long video benchmarks for two settings: 256-frame sampling with a 64-frame budget (256-64) and 128-frame sampling with a 32-frame budget (128-32). Our method outperforms baselines across all benchmarks and most subsets. Qwen2-VL-7B significantly outperforms baselines by up to 4.2\% on EgoSchema, 2.5\% on VideoMME, and 5.0\% on MLVU (256-64), and by up to 5.0\%, 3.8\%, and 5.4\% (128-32), with a 5.8\% gain on VideoMME medium subset. While our method also achieves notable improvements on LLaVA-OneVision-7B, the gains are less pronounced than on Qwen2-VL, likely due to noise from its one-dimensional positional encoding. The three-dimensional positional encoding of Qwen2-VL-7B offers more stable results, highlighting the importance of positional encoding design. Overall, these results demonstrate the effectiveness of our method and reveal some model-specific behaviors and limitations.

\item[]
\textbf{Short Video Results.} Short video benchmarks inherently contain fewer frames, simpler scenes, and primarily coherent motion, making them less affected by token length limitations. As a result, the trade-off between static and dynamic information is less pronounced, and performance gains tend to be smaller compared to long video settings. Nonetheless, we evaluate our method’s generalization on short video benchmarks by sampling 64 frames and setting the budget to 16 for videos averaging 10 seconds. As shown in Table~\ref{tab:short}, our method consistently outperforms all baselines on Qwen2-VL-7B, achieving up to 3.8\% higher accuracy and 0.16 higher scores. On LLaVA-OneVision-7B, it achieves strong results on ActivityNet-QA and performs comparably to the ``Retrieval'' baseline on MSVD-QA. These results demonstrate the robustness and generalization ability of our method even under short video scenarios.

\begin{table}[t]
    \centering
    \fontsize{9}{11}\selectfont
    \begin{tabular}{llcccc}
        \toprule
        \multirow{2}{*}{\textbf{Model}} & \multirow{2}{*}{\textbf{Method}} & \multicolumn{2}{c}{\textbf{MSVD-QA}} & \multicolumn{2}{c}{\makebox[0.6cm]{\textbf{ActivityNet-QA}}}\\
         & & Acc & Score & Acc & Score \\
        \midrule
        \multirow{4}{*}{Qwen2-VL} 
        & Original     & 66.0 & 3.59 & 50.3 & 2.82 \\
        & Retrieval    & 64.4 & 3.52 & 48.6 & 2.74 \\
        & Similarity   & 66.5 & 3.60 & 51.4	& 2.87 \\
        & \gc Ours     & \gc \tb{66.8} & \gc \tb{3.61} & \gc \tb{52.4} & \gc \tb{2.90} \\
        \midrule
        \multirow{4}{*}{\makecell[l]{LLaVA-\\OneVision}}
        & Original     & 54.3 & 3.09  & 52.6 & 2.90 \\
        & Retrieval    & \tb{54.8} & \tb{3.12} & 50.1 & 2.77  \\
        & Similarity   & 54.3 & 3.10 & 52.4 & 2.89 \\
        & \gc Ours     & \gc 54.7 & \gc 3.11  & \gc \tb{53.0} & \gc \tb{2.92} \\
        \bottomrule
    \end{tabular}
    \caption{Results on short video benchmarks. Although primarily focused on long videos, our method show stable and generalizable performance on short videos.}
    \label{tab:short}
\end{table}

\begin{figure*}[t]
  \includegraphics[width=\textwidth]{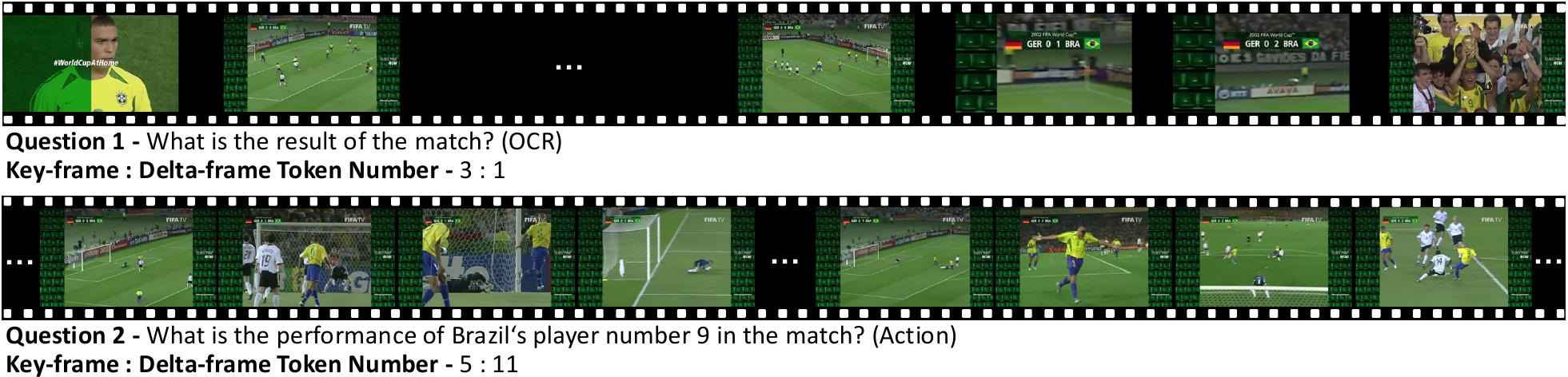}
  \caption{The qualitative analysis demonstrates that our method adjusts token allocation according to the query. For Question~1, an OCR task, the ratio of key-frame tokens to delta-frame tokens was 3:1. In contrast, for Question~2, an action recognition task, the ratio was 5:11.}
  \label{fig:case}
\end{figure*}

\begin{table}[t]
    \centering
    \fontsize{9}{11}\selectfont
    \begin{tabular}{cccc}
        \toprule
        \textbf{Method} & \makebox[1.2cm]{\textbf{EgoSchema}} & \makebox[1.2cm]{\textbf{VideoMME}} & \makebox[1.2cm]{\textbf{MLVU}}\\
        \midrule
        Original           & 64.7      & 57.6      & 46.8 \\
        Explore + Random  & 66.3      & 60.7      & 50.2 \\
        Explore + Select  & \tb{66.7} & \tb{61.4} & \tb{52.2}\\
        \bottomrule
    \end{tabular}
    \caption{Ablation study of our method. Results demonstrate the effectiveness of both stages, with each component yielding improvements over the baseline.}
    \label{tab:ablation}
\end{table}

\begin{table}[t]
    \centering
    \fontsize{9}{11}\selectfont
    \begin{tabular}{lcccc}
        \toprule
        \textbf{Model} & \textbf{Method} & \textbf{EgoSchema} & \textbf{VideoMME} \\
        \midrule
        \multirow{4}{*}{\makecell[l]{Qwen2-VL}} 
        & w/ query     & 66.3 & \tb{61.6}  \\
        & w/o query    & \tb{66.7} & 61.4  \\
        \cmidrule(lr){2-4}
        & mean     & 66.0 & 60.9 \\
        & max      & \tb{66.7} & \tb{61.4} \\
        \bottomrule
    \end{tabular}
    \caption{Ablation study on metric design. The first block shows that including the query token in $\bm{K}$ has a negligible impact, so it is omitted. The second block finds that the max operation in Equation~\eqref{eq:8} outperforms the mean on both benchmarks.}
    \label{tab:ablation_query}
\end{table}

\begin{figure}[t]
  \includegraphics[width=\columnwidth]{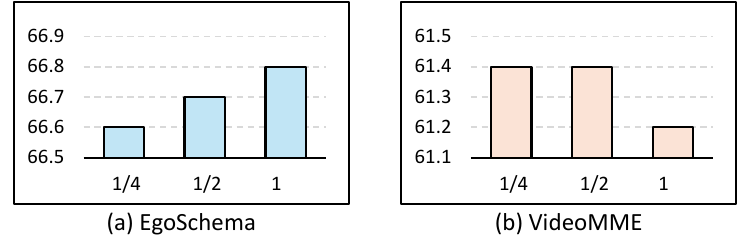}
  \caption{Search space size analysis. The x-axis represents the search space size. There are $n$ subsequences in the space, and their key frame number ranges from $\{1, 2, \dots, n\}$. Assuming the budget frame is $N_b$, ``$1$'' refers to $n = N_b$, ``$1/2$'' indicates $n = \lfloor N_b / 2 \rfloor$, ``$1/4$'' represents $n = \lfloor N_b / 4 \rfloor$. Larger search spaces benefit EgoSchema but hurt VideoMME and increase the time cost. A balanced setting uses half the budget size.}
  \label{fig:search}
\end{figure}

\end{itemize}

\subsection{Ablation Studies}

\begin{itemize}[leftmargin=0pt, itemindent=0pt]
\parindent 1em

\item[] \textbf{Stage Ablation.} As shown in Table~\ref{tab:ablation}, we conduct a two-stage ablation study on our method. The ablation experiments are performed on Qwen2-VL-7B, sampling 128 frames with a budget of 32 frames. First, we validate the effectiveness of the exploration stage. As indicated by the ``Explore + Random'' row in the table, generating multiple token subsequences followed by random selection results in improvement compared to the original operation, demonstrating the rationality of our search space design. Then we verify the effectiveness of the selection phase. On all benchmarks, our selection method achieves improvement over random selection.

\item[] \textbf{Metric Ablation.} Table~\ref{tab:ablation_query} presents two ablation studies on our metric design using Qwen2-VL-7B (128-frame sampling, 32-frame budget). The first block compares including or excluding the query token in the construction of $\bm{K}$ in Equation~\eqref{eq:6}, finding only marginal differences; for simplicity, we exclude the query token in our final design. The second block compares max and mean operations for query aggregation in Equation~\eqref{eq:8}, showing that the max operation consistently yields better results, thus supporting our metric choice.

\end{itemize}

\subsection{Further Analysis}

\begin{itemize}[leftmargin=0pt, itemindent=0pt]
\parindent 1em

\item[] \textbf{Qualitative Analysis.}
Figure~\ref{fig:case} shows two questions from the same video, both correctly answered using our method. In this example, we employ 128-frame sampling with a 32-frame budget, and the search space is defined as the full frame budget. Question 1, an OCR problem predominantly reliant on static information, prompts the method to allocate a key-to-delta-frame token ratio of 3:1. Conversely, action-related Question 2, necessitating the identification of a player scoring a goal, leads to the adoption of a key-to-delta-frame token ratio of 5:11.

\item[] \textbf{Search Space Size.}
We investigate the impact of search space size on performance using Qwen2-VL-7B, sampling 128 frames with a 32-frame budget. Figure~\ref{fig:search} shows that performance improves on EgoSchema when the search space matches the budget, but declines on VideoMME. We attribute this to excessive key frames, causing sparse delta-frame token selection and deviation from the training distribution, reducing effectiveness. Additionally, the time cost rises with the search space size. Therefore, we set the search space to half the frame budget to balance these factors.

\item[] \textbf{Efficiency Trade-off.}
With a limited token budget, sampling more frames prior to compression and input can improve performance. As shown in Table~\ref{tab:efficiency}, compressing 32 to 256 sampled frames into a 32-frame budget consistently improves accuracy. While compression introduces computational overhead, our primary focus is on memory efficiency and information retention. This approach involves a trade-off between performance and time cost; all reported time measurements are averaged over 10 runs to reduce variance. The main overhead stems from token selection and the increased load on the vision encoder. The selection cost, introduced by our framework, remains stable at approximately 0.5–0.6s and affects only the first-token latency, leaving subsequent decoding unaffected. The encoding cost scales roughly linearly with the number of sampled frames, but this is a general issue that can be largely mitigated by using asynchronous computation in multi-GPU environments.

\begin{table}[t]
    \centering
    \fontsize{9}{11}\selectfont
    \begin{tabular}{lccccc}
        \toprule
        \makebox[1.55cm][l]{\textbf{Metric}} & \tb{32} & \tb{64} & \tb{128} & \tb{256} & \tb{512} \\
        \midrule
        \makebox[1.55cm][l]{Accuracy (\%)}   & 46.8 & 52.0 & 52.2 & 54.0 & 54.6 \\
        \cmidrule(lr){1-6}
        \makebox[1.55cm][l]{Encoding (s)}    & 0.125 & 0.222 & 0.424 & 0.827 & 1.628 \\
        \makebox[1.55cm][l]{Selection (s)}   & -     & 0.537 & 0.557 & 0.565 & 0.575 \\
        \bottomrule
    \end{tabular}
    \caption{Performance and efficiency trade-off on the MLVU dataset using Qwen2-VL. Sampling more frames before compression improves accuracy. While the encoding cost increases roughly linearly (not specific to our method), the selection cost introduced by our framework remains stable.}
    \label{tab:efficiency}
\end{table}

\end{itemize}

\section{Conclusion}
Given that long videos possess tokens far exceeding the capacity that models can process, we advance token compression strategies by unveiling the following crucial fact: different question types exhibit varying dependencies on dynamic and static information. Based on this discovery, we propose a novel token selection strategy for visual token compression. Our method splits video frames into key and delta frames, and adaptively determines the optimal token allocations among key and delta frames guided by each specific query. Experiments demonstrate the effectiveness and generalizability of our method across multiple models and datasets.

\section*{Limitations}
In this paper, we propose a novel token selection strategy for visual token compression in video question answering, addressing the varying dependencies of questions on dynamic versus static video information. Our method has demonstrated effectiveness across multiple datasets, yet certain limitations remain. First, variations in positional encoding mechanisms across models may affect the ability to accurately estimate video length and temporally localize events. Nonetheless, we believe our approach offers valuable insights for designing compression modules in both pre-trained and fine-tuned video models. In terms of efficiency, our method introduces no additional memory overhead (superior to pruning in the key-value cache) but does incur extra time cost. This overhead mainly arises from the token selection process, where the selection metric is computed using the output of a shallow (second-layer) attention layer. Here, only the attention map between query and visual tokens is calculated, and multiple subsequences must be compared to finalize the selection. Importantly, this overhead occurs only during the initial token inference and does not affect subsequent decoding, and such costs are a common trade-off in most compression methods.


\section*{Acknowledgments}
This research is supported by the Ministry of Education, Singapore, under its Academic Research Fund Tier 1 (\#023618-00001, RG99/23), and the Start-Up Grant (\#023284-00001) of Nanyang Technological University, Singapore.

\bibliography{custom}

\begin{thebibliography}{40}
\providecommand{\natexlab}[1]{#1}

\bibitem[{Bai et~al.(2025)Bai, Chen, Liu, Wang, Ge, Song, Dang, Wang, Wang, Tang, Zhong, Zhu, Yang, Li, Wan, Wang, Ding, Fu, Xu, Ye, Zhang, Xie, Cheng, Zhang, Yang, Xu, and Lin}]{Qwen2.5-VL}
Shuai Bai, Keqin Chen, Xuejing Liu, Jialin Wang, Wenbin Ge, Sibo Song, Kai Dang, Peng Wang, Shijie Wang, Jun Tang, Humen Zhong, Yuanzhi Zhu, Mingkun Yang, Zhaohai Li, Jianqiang Wan, Pengfei Wang, Wei Ding, Zheren Fu, Yiheng Xu, Jiabo Ye, Xi~Zhang, Tianbao Xie, Zesen Cheng, Hang Zhang, Zhibo Yang, Haiyang Xu, and Junyang Lin. 2025.
\newblock Qwen2.5-vl technical report.
\newblock \emph{arXiv preprint arXiv:2502.13923}.

\bibitem[{Bolya et~al.(2022)Bolya, Fu, Dai, Zhang, Feichtenhofer, and Hoffman}]{bolya2022token}
Daniel Bolya, Cheng-Yang Fu, Xiaoliang Dai, Peizhao Zhang, Christoph Feichtenhofer, and Judy Hoffman. 2022.
\newblock Token merging: Your vit but faster.
\newblock \emph{arXiv preprint arXiv:2210.09461}.

\bibitem[{Chen et~al.(2024{\natexlab{a}})Chen, Zhao, Liu, Bai, Lin, Zhou, and Chang}]{chen2024image}
Liang Chen, Haozhe Zhao, Tianyu Liu, Shuai Bai, Junyang Lin, Chang Zhou, and Baobao Chang. 2024{\natexlab{a}}.
\newblock An image is worth 1/2 tokens after layer 2: Plug-and-play inference acceleration for large vision-language models.
\newblock In \emph{European Conference on Computer Vision}, pages 19--35. Springer.

\bibitem[{Chen et~al.(2024{\natexlab{b}})Chen, Wu, Wang, Su, Chen, Xing, Zhong, Zhang, Zhu, Lu et~al.}]{chen2024internvl}
Zhe Chen, Jiannan Wu, Wenhai Wang, Weijie Su, Guo Chen, Sen Xing, Muyan Zhong, Qinglong Zhang, Xizhou Zhu, Lewei Lu, et~al. 2024{\natexlab{b}}.
\newblock Internvl: Scaling up vision foundation models and aligning for generic visual-linguistic tasks.
\newblock In \emph{Proceedings of the IEEE/CVF Conference on Computer Vision and Pattern Recognition}, pages 24185--24198.

\bibitem[{Choudhury et~al.(2024)Choudhury, Zhu, Liu, Niinuma, Kitani, and Jeni}]{choudhury2024don}
Rohan Choudhury, Guanglei Zhu, Sihan Liu, Koichiro Niinuma, Kris~M Kitani, and L{\'a}szl{\'o} Jeni. 2024.
\newblock Don't look twice: Faster video transformers with run-length tokenization.
\newblock \emph{arXiv preprint arXiv:2411.05222}.

\bibitem[{Fu et~al.(2024{\natexlab{a}})Fu, Dai, Luo, Li, Ren, Zhang, Wang, Zhou, Shen, Zhang et~al.}]{fu2024video}
Chaoyou Fu, Yuhan Dai, Yongdong Luo, Lei Li, Shuhuai Ren, Renrui Zhang, Zihan Wang, Chenyu Zhou, Yunhang Shen, Mengdan Zhang, et~al. 2024{\natexlab{a}}.
\newblock Video-mme: The first-ever comprehensive evaluation benchmark of multi-modal llms in video analysis.
\newblock \emph{arXiv preprint arXiv:2405.21075}.

\bibitem[{Fu et~al.(2024{\natexlab{b}})Fu, Liu, Han, Dai, Yan, Yang, Ning, and Wang}]{fu2024framefusion}
Tianyu Fu, Tengxuan Liu, Qinghao Han, Guohao Dai, Shengen Yan, Huazhong Yang, Xuefei Ning, and Yu~Wang. 2024{\natexlab{b}}.
\newblock Framefusion: Combining similarity and importance for video token reduction on large visual language models.
\newblock \emph{arXiv preprint arXiv:2501.01986}.

\bibitem[{He et~al.(2024{\natexlab{a}})He, Li, Jang, Jia, Cao, Shah, Shrivastava, and Lim}]{he2024ma}
Bo~He, Hengduo Li, Young~Kyun Jang, Menglin Jia, Xuefei Cao, Ashish Shah, Abhinav Shrivastava, and Ser-Nam Lim. 2024{\natexlab{a}}.
\newblock Ma-lmm: Memory-augmented large multimodal model for long-term video understanding.
\newblock In \emph{Proceedings of the IEEE/CVF Conference on Computer Vision and Pattern Recognition}, pages 13504--13514.

\bibitem[{He et~al.(2024{\natexlab{b}})He, Chen, Liu, Shao, Zhou, Zhang, and Zhuang}]{he2024zipvl}
Yefei He, Feng Chen, Jing Liu, Wenqi Shao, Hong Zhou, Kaipeng Zhang, and Bohan Zhuang. 2024{\natexlab{b}}.
\newblock Zipvl: Efficient large vision-language models with dynamic token sparsification and kv cache compression.
\newblock \emph{arXiv preprint arXiv:2410.08584}.

\bibitem[{Lan et~al.(2024)Lan, Yuan, Jie, and Ma}]{lan2024vidcompress}
Xiaohan Lan, Yitian Yuan, Zequn Jie, and Lin Ma. 2024.
\newblock Vidcompress: Memory-enhanced temporal compression for video understanding in large language models.
\newblock \emph{arXiv preprint arXiv:2410.11417}.

\bibitem[{Lee et~al.(2006)Lee, Shin, and Park}]{lee2006adaptive}
Jeehong Lee, IlHong Shin, and HyunWook Park. 2006.
\newblock Adaptive intra-frame assignment and bit-rate estimation for variable gop length in h. 264.
\newblock \emph{IEEE Transactions on Circuits and Systems for Video Technology}, 16(10):1271--1279.

\bibitem[{Li et~al.(2024{\natexlab{a}})Li, Zhang, Guo, Zhang, Li, Zhang, Zhang, Zhang, Li, Liu et~al.}]{li2024llava}
Bo~Li, Yuanhan Zhang, Dong Guo, Renrui Zhang, Feng Li, Hao Zhang, Kaichen Zhang, Peiyuan Zhang, Yanwei Li, Ziwei Liu, et~al. 2024{\natexlab{a}}.
\newblock Llava-onevision: Easy visual task transfer.
\newblock \emph{arXiv preprint arXiv:2408.03326}.

\bibitem[{Li et~al.(2023{\natexlab{a}})Li, Li, Savarese, and Hoi}]{li2023blip}
Junnan Li, Dongxu Li, Silvio Savarese, and Steven Hoi. 2023{\natexlab{a}}.
\newblock Blip-2: Bootstrapping language-image pre-training with frozen image encoders and large language models.
\newblock In \emph{International conference on machine learning}, pages 19730--19742. PMLR.

\bibitem[{Li et~al.(2023{\natexlab{b}})Li, He, Wang, Li, Wang, Luo, Wang, Wang, and Qiao}]{li2023videochat}
KunChang Li, Yinan He, Yi~Wang, Yizhuo Li, Wenhai Wang, Ping Luo, Yali Wang, Limin Wang, and Yu~Qiao. 2023{\natexlab{b}}.
\newblock Videochat: Chat-centric video understanding.
\newblock \emph{arXiv preprint arXiv:2305.06355}.

\bibitem[{Li et~al.(2024{\natexlab{b}})Li, Yuan, Liu, Tang, Wang, Qin, Zhu, and Zhang}]{li2024tokenpacker}
Wentong Li, Yuqian Yuan, Jian Liu, Dongqi Tang, Song Wang, Jie Qin, Jianke Zhu, and Lei Zhang. 2024{\natexlab{b}}.
\newblock Tokenpacker: Efficient visual projector for multimodal llm.
\newblock \emph{arXiv preprint arXiv:2407.02392}.

\bibitem[{Li et~al.(2024{\natexlab{c}})Li, Wang, and Jia}]{li2024llama}
Yanwei Li, Chengyao Wang, and Jiaya Jia. 2024{\natexlab{c}}.
\newblock Llama-vid: An image is worth 2 tokens in large language models.
\newblock In \emph{European Conference on Computer Vision}, pages 323--340. Springer.

\bibitem[{Lin et~al.(2023)Lin, Zhu, Ye, Ning, Jin, and Yuan}]{lin2023video}
Bin Lin, Bin Zhu, Yang Ye, Munan Ning, Peng Jin, and Li~Yuan. 2023.
\newblock Video-llava: Learning united visual representation by alignment before projection.
\newblock \emph{arXiv preprint arXiv:2311.10122}.

\bibitem[{Liu et~al.(2024)Liu, Wang, Ma, Wu, Ma, Wei, Jiao, Wu, and Hu}]{kangaroogroup}
Jiajun Liu, Yibing Wang, Hanghang Ma, Xiaoping Wu, Xiaoqi Ma, xiaoming Wei, Jianbin Jiao, Enhua Wu, and Jie Hu. 2024.
\newblock Kangaroo: A powerful video-language model supporting long-context video input.
\newblock \emph{arXiv preprint arXiv:2408.15542}.

\bibitem[{Mangalam et~al.(2023)Mangalam, Akshulakov, and Malik}]{mangalam2023egoschema}
Karttikeya Mangalam, Raiymbek Akshulakov, and Jitendra Malik. 2023.
\newblock Egoschema: A diagnostic benchmark for very long-form video language understanding.
\newblock \emph{Advances in Neural Information Processing Systems}, 36:46212--46244.

\bibitem[{Mogrovejo and Solorio(2024)}]{mogrovejo2024question}
David Mogrovejo and Thamar Solorio. 2024.
\newblock Question-instructed visual descriptions for zero-shot video answering.
\newblock In \emph{Findings of the Association for Computational Linguistics ACL 2024}, pages 9329--9339.

\bibitem[{Nie et~al.(2024)Nie, Ding, Wang, Guo, Han, Xu, and Zhang}]{nie2024slowfocus}
Ming Nie, Dan Ding, Chunwei Wang, Yuanfan Guo, Jianhua Han, Hang Xu, and Li~Zhang. 2024.
\newblock Slowfocus: Enhancing fine-grained temporal understanding in video llm.
\newblock In \emph{The Thirty-eighth Annual Conference on Neural Information Processing Systems}.

\bibitem[{OpenAI(2024)}]{openai2024gpt4o}
OpenAI. 2024.
\newblock Gpt-4o system card.
\newblock \url{https://openai.com/index/gpt-4o-system-card/}.

\bibitem[{Oquab et~al.(2023)Oquab, Darcet, Moutakanni, Vo, Szafraniec, Khalidov, Fernandez, Haziza, Massa, El-Nouby et~al.}]{oquab2023dinov2}
Maxime Oquab, Timoth{\'e}e Darcet, Th{\'e}o Moutakanni, Huy Vo, Marc Szafraniec, Vasil Khalidov, Pierre Fernandez, Daniel Haziza, Francisco Massa, Alaaeldin El-Nouby, et~al. 2023.
\newblock Dinov2: Learning robust visual features without supervision.
\newblock \emph{arXiv preprint arXiv:2304.07193}.

\bibitem[{Qian et~al.(2025)Qian, Dong, Zhang, Zang, Ding, Lin, and Wang}]{qian2025streaming}
Rui Qian, Xiaoyi Dong, Pan Zhang, Yuhang Zang, Shuangrui Ding, Dahua Lin, and Jiaqi Wang. 2025.
\newblock Streaming long video understanding with large language models.
\newblock \emph{Advances in Neural Information Processing Systems}, 37:119336--119360.

\bibitem[{Shen et~al.(2024)Shen, Xiong, Zhao, Wu, Chen, Zhu, Liu, Xiao, Varadarajan, Bordes, Liu, Xu, J.~Kim, Soran, Krishnamoorthi, Elhoseiny, and Chandra}]{shen2024longvu}
Xiaoqian Shen, Yunyang Xiong, Changsheng Zhao, Lemeng Wu, Jun Chen, Chenchen Zhu, Zechun Liu, Fanyi Xiao, Balakrishnan Varadarajan, Florian Bordes, Zhuang Liu, Hu~Xu, Hyunwoo J.~Kim, Bilge Soran, Raghuraman Krishnamoorthi, Mohamed Elhoseiny, and Vikas Chandra. 2024.
\newblock Longvu: Spatiotemporal adaptive compression for long video-language understanding.
\newblock \emph{arXiv preprint arXiv:2410.17434}.

\bibitem[{Song et~al.(2024)Song, Chai, Wang, Zhang, Zhou, Wu, Chi, Guo, Ye, Zhang et~al.}]{song2024moviechat}
Enxin Song, Wenhao Chai, Guanhong Wang, Yucheng Zhang, Haoyang Zhou, Feiyang Wu, Haozhe Chi, Xun Guo, Tian Ye, Yanting Zhang, et~al. 2024.
\newblock Moviechat: From dense token to sparse memory for long video understanding.
\newblock In \emph{Proceedings of the IEEE/CVF Conference on Computer Vision and Pattern Recognition}, pages 18221--18232.

\bibitem[{Team et~al.(2024)Team, Georgiev, Lei, Burnell, Bai, Gulati, Tanzer, Vincent, Pan, Wang et~al.}]{team2024gemini}
Gemini Team, Petko Georgiev, Ving~Ian Lei, Ryan Burnell, Libin Bai, Anmol Gulati, Garrett Tanzer, Damien Vincent, Zhufeng Pan, Shibo Wang, et~al. 2024.
\newblock Gemini 1.5: Unlocking multimodal understanding across millions of tokens of context.
\newblock \emph{arXiv preprint arXiv:2403.05530}.

\bibitem[{Wan et~al.(2024)Wan, Wu, Liu, Huang, Zhu, Jin, Wang, and Yuan}]{wan2024look}
Zhongwei Wan, Ziang Wu, Che Liu, Jinfa Huang, Zhihong Zhu, Peng Jin, Longyue Wang, and Li~Yuan. 2024.
\newblock Look-m: Look-once optimization in kv cache for efficient multimodal long-context inference.
\newblock \emph{arXiv preprint arXiv:2406.18139}.

\bibitem[{Wang et~al.(2024)Wang, Bai, Tan, Wang, Fan, Bai, Chen, Liu, Wang, Ge et~al.}]{wang2024qwen2}
Peng Wang, Shuai Bai, Sinan Tan, Shijie Wang, Zhihao Fan, Jinze Bai, Keqin Chen, Xuejing Liu, Jialin Wang, Wenbin Ge, et~al. 2024.
\newblock Qwen2-vl: Enhancing vision-language model's perception of the world at any resolution.
\newblock \emph{arXiv preprint arXiv:2409.12191}.

\bibitem[{Xu et~al.(2017)Xu, Zhao, Xiao, Wu, Zhang, He, and Zhuang}]{xu2017video}
Dejing Xu, Zhou Zhao, Jun Xiao, Fei Wu, Hanwang Zhang, Xiangnan He, and Yueting Zhuang. 2017.
\newblock Video question answering via gradually refined attention over appearance and motion.
\newblock In \emph{Proceedings of the 25th ACM international conference on Multimedia}, pages 1645--1653.

\bibitem[{Yao et~al.(2024)Yao, Yu, Zhang, Wang, Cui, Zhu, Cai, Li, Zhao, He et~al.}]{yao2024minicpm}
Yuan Yao, Tianyu Yu, Ao~Zhang, Chongyi Wang, Junbo Cui, Hongji Zhu, Tianchi Cai, Haoyu Li, Weilin Zhao, Zhihui He, et~al. 2024.
\newblock Minicpm-v: A gpt-4v level mllm on your phone.
\newblock \emph{arXiv preprint arXiv:2408.01800}.

\bibitem[{Ye et~al.(2024)Ye, Xu, Liu, Hu, Yan, Qian, Zhang, Huang, and Zhou}]{ye2024mplug}
Jiabo Ye, Haiyang Xu, Haowei Liu, Anwen Hu, Ming Yan, Qi~Qian, Ji~Zhang, Fei Huang, and Jingren Zhou. 2024.
\newblock mplug-owl3: Towards long image-sequence understanding in multi-modal large language models.
\newblock \emph{arXiv preprint arXiv:2408.04840}.

\bibitem[{Ye et~al.(2023)Ye, Xu, Xu, Ye, Yan, Zhou, Wang, Hu, Shi, Shi et~al.}]{ye2023mplug}
Qinghao Ye, Haiyang Xu, Guohai Xu, Jiabo Ye, Ming Yan, Yiyang Zhou, Junyang Wang, Anwen Hu, Pengcheng Shi, Yaya Shi, et~al. 2023.
\newblock mplug-owl: Modularization empowers large language models with multimodality.
\newblock \emph{arXiv preprint arXiv:2304.14178}.

\bibitem[{Yu et~al.(2019)Yu, Xu, Yu, Yu, Zhao, Zhuang, and Tao}]{yu2019activitynet}
Zhou Yu, Dejing Xu, Jun Yu, Ting Yu, Zhou Zhao, Yueting Zhuang, and Dacheng Tao. 2019.
\newblock Activitynet-qa: A dataset for understanding complex web videos via question answering.
\newblock In \emph{Proceedings of the AAAI Conference on Artificial Intelligence}, volume~33, pages 9127--9134.

\bibitem[{Zhang et~al.(2025)Zhang, Li, Cheng, Hu, Yuan, Chen, Leng, Jiang, Zhang, Li et~al.}]{damonlpsg2025videollama3}
Boqiang Zhang, Kehan Li, Zesen Cheng, Zhiqiang Hu, Yuqian Yuan, Guanzheng Chen, Sicong Leng, Yuming Jiang, Hang Zhang, Xin Li, et~al. 2025.
\newblock \href {https://arxiv.org/abs/2501.13106} {Videollama 3: Frontier multimodal foundation models for image and video understanding}.
\newblock \emph{arXiv preprint arXiv:2501.13106}.

\bibitem[{Zhang et~al.(2024{\natexlab{a}})Zhang, Lu, Islam, Wang, Yu, Bansal, and Bertasius}]{zhang-etal-2024-simple}
Ce~Zhang, Taixi Lu, Md~Mohaiminul Islam, Ziyang Wang, Shoubin Yu, Mohit Bansal, and Gedas Bertasius. 2024{\natexlab{a}}.
\newblock \href {https://doi.org/10.18653/v1/2024.emnlp-main.1209} {A simple {LLM} framework for long-range video question-answering}.
\newblock In \emph{Proceedings of the 2024 Conference on Empirical Methods in Natural Language Processing}, pages 21715--21737, Miami, Florida, USA. Association for Computational Linguistics.

\bibitem[{Zhang et~al.(2024{\natexlab{b}})Zhang, Zhang, Li, Zeng, Yang, Zhang, Wang, Tan, Li, and Liu}]{zhang2024longva}
Peiyuan Zhang, Kaichen Zhang, Bo~Li, Guangtao Zeng, Jingkang Yang, Yuanhan Zhang, Ziyue Wang, Haoran Tan, Chunyuan Li, and Ziwei Liu. 2024{\natexlab{b}}.
\newblock \href {https://arxiv.org/abs/2406.16852} {Long context transfer from language to vision}.
\newblock \emph{arXiv preprint arXiv:2406.16852}.

\bibitem[{Zhang et~al.(2024{\natexlab{c}})Zhang, Wu, Li, Li, Ma, Liu, and Li}]{zhang2024video}
Yuanhan Zhang, Jinming Wu, Wei Li, Bo~Li, Zejun Ma, Ziwei Liu, and Chunyuan Li. 2024{\natexlab{c}}.
\newblock Video instruction tuning with synthetic data.
\newblock \emph{arXiv preprint arXiv:2410.02713}.

\bibitem[{Zhang et~al.(2023)Zhang, Sheng, Zhou, Chen, Zheng, Cai, Song, Tian, R{\'e}, Barrett et~al.}]{zhang2023h2o}
Zhenyu Zhang, Ying Sheng, Tianyi Zhou, Tianlong Chen, Lianmin Zheng, Ruisi Cai, Zhao Song, Yuandong Tian, Christopher R{\'e}, Clark Barrett, et~al. 2023.
\newblock H2o: Heavy-hitter oracle for efficient generative inference of large language models.
\newblock \emph{Advances in Neural Information Processing Systems}, 36:34661--34710.

\bibitem[{Zhou et~al.(2024)Zhou, Shu, Zhao, Wu, Xiao, Yang, Xiong, Zhang, Huang, and Liu}]{MLVU}
Junjie Zhou, Yan Shu, Bo~Zhao, Boya Wu, Shitao Xiao, Xi~Yang, Yongping Xiong, Bo~Zhang, Tiejun Huang, and Zheng Liu. 2024.
\newblock Mlvu: A comprehensive benchmark for multi-task long video understanding.
\newblock \emph{arXiv preprint arXiv:2406.04264}.

\end{thebibliography}

\appendix

\section{Additional Experiments}
\label{sec:appendix_exp}
The appendix presents additional experiments, including results on the advanced Qwen2.5-VL model and comparisons between our method and the reproduced training-free LongVU approach.

\subsection{Experiments on Qwen2.5-VL}
\label{appendix_qwen2.5}
Qwen2.5-VL is the latest vision language model in the Qwen series models, officially released in February 2025. Building upon the foundation of Qwen2-VL, Qwen2.5-VL introduces significant enhancements in long video comprehension. Notably, it incorporates absolute time encoding, enabling the model to handle videos of extended durations with second-level event localization. To provide a more comprehensive evaluation of our method, we report experimental results on the Qwen2.5-VL-7B model using the same experimental settings as in the main text.

\begin{table*}[!h]
    \centering
    \fontsize{9}{11}\selectfont
    \begin{tabular}{llcccccccc}
        \toprule
        \multirow{2}{*}{\textbf{Model}} & \multicolumn{3}{c}{\textbf{Settings}} & \textbf{EgoSchema} & \multicolumn{4}{c}{\textbf{VideoMME}} & \textbf{MLVU} \\
        & Method & Sample & Budget & & Short & Medium & Long & Overall \\
        \midrule
        \multirow{8}{*}{Qwen2.5-VL-7B} 
        & Original   & 256 & -  & 60.3 & 75.0 & 61.8 & 51.0 & 62.6 & 50.0 \\
        & Retrieval  & 256 & 64 & 60.9 & 75.4 & \textbf{66.7} & 54.8 & 65.6 & 56.2 \\
        & Similarity & 256 & 64 & 60.8 & 74.0 & 64.7 & 54.3 & 64.3 & 53.6 \\
        & \gc Ours   & \gc 256 & \gc 64 & \gc \tb{61.6} & \gc \tb{75.8} & \gc 65.2 & \gc \tb{56.1} & \gc \tb{65.7} & \gc \tb{58.4} \\
        \cmidrule(lr){2-10}
        & Original   & 128 & -  & 59.1 & 73.1 & 60.0 & 49.6 & 60.9 & 47.2 \\
        & Retrieval  & 128 & 32 & 60.2 & \tb{74.6} & \tb{64.8} & 53.3 & \tb{64.2} & 48.4 \\
        & Similarity & 128 & 32 & 60.0 & 73.3 & 60.9 & 51.6 & 61.9 & 47.6 \\
        & \gc Ours   & \gc 128 & \gc 32 & \gc \tb{60.6} & \gc{74.1} & \gc{63.2} & \gc \tb{53.9} & \gc{63.7} & \gc \tb{51.6} \\
        \bottomrule
    \end{tabular}
    \caption{Long video benchmark results on Qwen2.5-VL-7B. Our method achieves the best performance on EgoSchema, MLVU, and the VideoMME long subset, with improvements of up to 5.1\% on the VideoMME long subset and 8.4\% on MLVU over the baselines, demonstrating strong effectiveness and generalization.}
    \label{tab:long_qwen2.5}
\end{table*}

\begin{table}[!htb]
    \centering
    \fontsize{9}{11}\selectfont
    \begin{tabular}{llcc}
        \toprule
        \multirow{2}{*}{\textbf{Model}} & \multirow{2}{*}{\textbf{Method}} & \multicolumn{2}{c}{\textbf{ActivityNet-QA}}\\
         & & \hspace{0.1cm} Accuracy \hspace{0.1cm} & \hspace{0.1cm} Score \hspace{0.1cm} \\
        \midrule
        \multirow{4}{*}{Qwen2.5-VL} 
        & Original     & 52.1 & 2.96 \\
        & Retrieval    & 52.7 & 2.98 \\
        & Similarity   & 53.1 & 2.99 \\
        & \gc Ours     & \gc \tb{54.3} & \gc \tb{3.07} \\
        \bottomrule
    \end{tabular}
    \caption{Short video benchmark results on Qwen2.5-VL. Our method achieves the highest accuracy and score, outperforming all baselines.}
    \label{tab:short_qwen2.5}
\end{table}

\begin{table}[!h]
    \centering
    \fontsize{9}{11}\selectfont
    \begin{tabular}{llcc}
        \toprule
        \textbf{Model} & \textbf{Method} & \textbf{EgoSchema} & \textbf{VideoMME} \\
        \midrule
        \multirow{3}{*}{Qwen2-VL} 
        & Original     & 66.2 & 60.4  \\
        & LongVU       & 67.2 & 62.3  \\
        & \gc Ours     & \gc \tb{67.8} & \gc \tb{62.9} \\
        \midrule
        \multirow{3}{*}{\makecell[l]{LLaVA-\\OneVision}} 
        & Original     & 60.1 & 58.0  \\
        & LongVU       & \tb{60.3} & 59.3  \\
        & \gc Ours     & \gc \tb{60.3} & \gc \tb{60.6}  \\
        \midrule
        \multirow{3}{*}{Qwen2.5-VL} 
        & Original     & 60.3 & 62.6 \\
        & LongVU       & 61.6 & 64.5 \\
        & \gc Ours     & \gc \tb{61.6} & \gc \tb{65.7} \\
        \bottomrule
    \end{tabular}
    \caption{Comparison with the reproduced training-free LongVU. Our method consistently outperforms the reproduced LongVU across models and benchmarks, while offering more precise control over the token count.}
    \label{tab:longvu_combined}
\end{table}

\begin{itemize}[leftmargin=0pt, itemindent=0pt]
\parindent 1em

\item[] \textbf{Long Video Results.} Table~\ref{tab:long_qwen2.5} presents the long video benchmark results on Qwen2.5-VL-7B under different sampling and budget settings. Across both the 256-64 and 128-32 configurations, our method consistently achieves the best performance on EgoSchema, MLVU, and the VideoMME long subset. Notably, it improves performance on the VideoMME long subset and MLVU by up to 5.1\% and 8.4\% compared to the baselines. For the VideoMME short and medium subsets, due to the shorter video length, our approach does not significantly outperform all baselines but still delivers strong results. These findings demonstrate the effectiveness and robustness of our method and further validate its strong generalization capability across different models.

\item[] \textbf{Short Video Results.} 
Although our method is primarily designed for long video understanding, it also delivers strong results on short video tasks. For example, on Qwen2.5-VL evaluated with ActivityNet-QA under the 64-frame sampling and 16-frame budget setting, our approach achieves the best performance among all baselines. As shown in Table~\ref{tab:short_qwen2.5}, it attains the highest accuracy of 54.3\% and a score of 3.07, outperforming the baselines by up to 2.2\% in accuracy and 0.11 in score. These results further demonstrate that our method remains robust and effective across different video lengths.
\end{itemize}

\subsection{Comparison with Training-free LongVU}
\label{app_longvu}
To further highlight the advantages of our approach, we compare it with LongVU~\cite{shen2024longvu} by reproducing its compression method in a training-free setting under the 256-frame sampling and 64-frame budget configuration. Following the original paper, we use DINOv2~\cite{oquab2023dinov2} with a 0.83 threshold for frame reduction and apply a $\lfloor2/3\rfloor$ downsampling ratio. However, we observe that achieving an exact token budget with LongVU requires careful tuning of thresholds and heuristics, providing only indirect control over compression. In contrast, our method employs top-K selection, allowing direct and precise control of the token count. As shown in Table~\ref{tab:longvu_combined}, our approach consistently outperforms the reproduced LongVU across all models and benchmarks, while offering more reliable and practical token budget management.

\section{Prompt Details}
\label{sec:appendix}
We use the template provided by the model for the instruction prompt, while introducing our textual organization format only in the questioning part.

\subsection{Prompts for Multiple-Choice Questions}
We add the sentence "Respond with only the letter (A, B, C, or D) of the correct option." at the beginning of the multiple-choice questions. Here is an example for questions in VideoMME: \\
\noindent\hrule
\vspace{0.2cm}

\noindent \textit{Respond with only the letter (A, B, C, or D) of the correct option. \\
Which elements are depicted in the painting introduced by the video?\\
A. A little girl and a red balloon.\\
B. A little boy and a red balloon.\\
C. A little girl and a blue balloon.\\
D. An adult and a blue balloon.
}
\vspace{0.2cm}
\noindent\hrule 
\vspace{0.6cm}

Here is an example for EgoSchema: \\
\noindent\hrule
\vspace{0.2cm}

\noindent \textit{Respond with only the letter (A, B, C, D or E) of the correct option. \\
Taking into account all the actions performed by c, what can you deduce about the primary objective and focus within the video content? \\
A. C is cooking. \\
B. C is doing laundry. \\
C. C is cleaning the kitchen. \\
D. C is cleaning dishes. \\
E. C is cleaning the bathroom.
}
\vspace{0.2cm}
\noindent\hrule
\vspace{0.6cm}

And here is an example for MLVU: \\
\noindent\hrule
\vspace{0.2cm}

\noindent \textit{Respond with only the letter (A, B, C, D, E or F) of the correct option. \\
In what setting does the video take place? \\
(A) Castle \\
(B) Forest \\
(C) Desert \\
(D) Countryside \\
(E) Ocean \\
(F) Campus
}
\vspace{0.2cm}
\noindent\hrule
\vspace{0.5cm}

\subsection{Prompts for Open-Ended Questions}
We add the sentence "Answer the question according to the video." at the beginning of the open-ended questions. Here is an example: \\
\noindent\hrule
\vspace{0.2cm}

\noindent \textit{Answer the question according to the video. \\
Who did circles on the back tire of his motorcycle in the parking lot?
}
\vspace{0.2cm}
\noindent\hrule



\end{document}